%% file: elsarticle-template-num.tex
\documentclass[authoryear,preprint,12pt]{elsarticle}

\usepackage{amssymb}
\usepackage{tabularx}
\usepackage{multirow}
\usepackage{diagbox}
\usepackage{stfloats}
\usepackage[dvipsnames]{xcolor}
\usepackage{graphicx} %package to manage images
\graphicspath{ {image/} }

\usepackage[english]{babel}
\usepackage[utf8]{inputenc}
\usepackage[ruled, linesnumbered]{algorithm2e}
\usepackage{url}
\newcommand{\txtRed}[1]{\textcolor{black}{#1}}
\journal{Computers in Industry}

\begin{document}
\begin{frontmatter}



\title{Component Segmentation of Engineering Drawings Using Graph Convolutional Networks
}


\author{Wentai Zhang}
\author{Joe Joseph}
\author{Yue Yin}
\author{Liuyue Xie}
\author{Tomotake Furuhata}
\author{Soji Yamakawa}
\author{Kenji Shimada}
\author{Levent Burak Kara\corref{cor1}}

% \fntext[cor2]{Equal contribution.}
\cortext[cor1]{Address all correspondences to \ead{lkara@cmu.edu}}

\address{Department of Mechanical Engineering, Carnegie Mellon University, Pittsburgh, PA, 15213, USA}
%%%%%%%%%%%%%%%%%%%%%%%%%%%%%%%%%%%%%%%%%%%%%%%%%%%%%%%%%%%%%%
\begin{abstract}
We present a data-driven framework to automate the vectorization and machine interpretation of 2D engineering part drawings. In industrial settings, most manufacturing engineers still rely on manual reads to identify the topological and manufacturing requirements from drawings submitted by designers. The interpretation process is laborious and time-consuming, which severely inhibits the efficiency of part quotation and manufacturing tasks. While recent advances in image-based computer vision methods have demonstrated great potential in interpreting natural images through semantic segmentation approaches, the application of such methods in parsing engineering technical drawings into semantically accurate components remains a significant challenge. The severe pixel sparsity in engineering drawings also restricts the effective featurization of image-based data-driven methods. To overcome these challenges, we propose a deep learning based framework that predicts the semantic type of each vectorized component. Taking a raster image as input, we vectorize all components through thinning, stroke tracing, and cubic bezier fitting. Then a graph of such components is generated based on the connectivity between the components. Finally, a graph convolutional neural network is trained on this graph data to identify the semantic type of each component. We test our framework in the context of semantic segmentation of text, dimension and, contour components in engineering drawings. Results show that our method yields the best performance compared to recent image-based, and graph-based segmentation methods.

% We present a new data generation method to facilitate an automatic machine interpretation of 2D engineering part drawings. While such drawings are a very common medium for clients to encode design and manufacturing requirements, a lack of computer support to automatically interpret these drawings necessitates part manufacturers to resort to laborious manual approaches for interpretation which, in turn, severely limits processing capacity.  Although recent advances in trainable computer vision methods may enable automatic machine interpretation, it remains challenging to apply such methods to engineering drawings due to a lack of labelled training data. As one step toward this challenge, we propose a constrained data synthesis method to  generate an arbitrarily large set of synthetic training drawings using only a handful of labelled examples. Our method is based on the randomization of the dimension sets subject to two major constraints to ensure the validity of the synthetic drawings. The effectiveness of our method is demonstrated in the context of a  binary component segmentation task with a proposed list of descriptors. An evaluation of several image segmentation methods trained on our synthetic dataset shows that our approach to new data generation can boost the segmentation accuracy and the generalizability of the machine learning models to unseen drawings. 
\end{abstract}

\begin{keyword}
engineering drawings \sep graph neural networks \sep component segmentation \sep deep learning \sep computer vision

%% or \MSC[2008] code \sep code  (2000 is the default)
% point cloud, upsampling, deep learning, neural network
\end{keyword}
%%%%%%%%%%%%%%%%%%%%%%%%%%%%%%%%%%%%%%%%%%%%%%%%%%%%%%%%%%%%%%%
\end{frontmatter}



% \begin{figure*}[!htp]
%   \centering
%   \includegraphics[trim = 0in 0in 0in 0in, clip, width=\textwidth]{pipeline.png}
%   \caption{A schematic graph of our proposed pipeline. Taking part drawing images as input, our pipeline is able to perform the component segmentation with vectorization, graph extraction and classifier training. }
%   \label{fig:pipeline}
% \end{figure*}

\section{Introduction} \label{intro}
\input{tex/1Introduction}

\input{tex/Contributions}
%%%%%%%%%%%%%%%%%%%%%%%%%%%%%%%%%%%%%%%%%%%%%%%%%%%%%%%%%%%%%%%%%%%%%%%%%%%%%%%%%%%%%%%%%%%%%%%%%%%%%%%%
\section{Related Works}

\input{tex/2Relatedworks}

%%%%%%%%%%%%%%%%%%%%%%%%%%%%%%%%%%%%%%%%%%%%%%%%%%%%%%%%%%%%%%%%%%%%%%
\section{Technical Approach}

%%%%%%%%%%%%%%%%%%%%%%%% 

\input{tex/3techical_approach}

% An algorithm to embed the information in engineering drawing bitmaps into a sparsely connected graph that indicates all important components in the graph and their relationships.
% constraints on the dimension sets generation
% Vectorization of the bitmaps + island formation
% Definition of the visual features. (Nodal features)
% Construction of the edges between the nodes. (Edge)

% A deep learning pipeline that takes the obtained unlabelled graphs as input and predicts the component type(contour/dimension) for each node in the graph.
% The generation of the synthetic dataset: 
% Random dimension sets generation
% Strategies towards a more practical outcome (avoid overlapping, control the number)
% The choice of network (GCN, GraphSAGE, GAT)
% Network architecture, training process, hyperparameter study,

% \begin{figure}[!htp]
%   \centering
%   \includegraphics[trim = 0in 0in 0in 0in, clip, width=\columnwidth]{sampleX.png}
%   \caption{Example results of the two subsampling strategies for the input point clouds.  (a) Ground truth point cloud.  (b) Uniform subsampling.  (c) Curvature-based subsampling.}
%   \label{fig:sample}
% \end{figure}

%%%%%%%%%%%%%%%%%%%%%%%%%%%%%%%%%%%%%%%%%%%%%%%%%%%%%%%%%%%%%%%%%%%%%%%
% \section{Experiments}

%%%%%%%%%%%%%%%%%%%%%%%%%%%%%%%%%%%%%%%%%%%%%%%%%%%%%%%%%%%%%%%%%%%%%%%
\section{Results}\label{chp:cs_res}

\input{tex/4results}

%%%%%%%%%%%%%%%%%%%%%%%%%%%%%%%%%%%%%%%%%%%%%%%%%%%%%%%%%%%%%%%%%%%%%%%%
\section{Discussions and Future Work}

\input{tex/5future}

%%%%%%%%%%%%%%%%%%%%%%%%%%%%%%%%%%%%%%%%%%%%%%%%%%%%%%%%%%%%%%%%%%%%%%%%
\section{Conclusions}

\input{tex/6conclusion}

%%%%%%%%%%%%%%%%%%%%%%%%%%%%%%%%%%%%%%%%%%%%%%%%%%%%%%%%%%%%%%%%%%%%%%%%
\section*{Acknowledgement}
The authors would like to thank MiSUMi Corporation for their provision of a contemporary engineering problem, guidance on the applicability of developed methods, and financial support. Additionally, the authors appreciate the help from Run Wang, Zhuoran Cheng, Zheren Zhu, and Chenlai Wang.
%%%%%%%%%%%%%%%%%%%%%%%%%%%%%%%%%%%%%%%%%%%%%%%%%%%%%%%%%%%%%%%%%%%%%%%

\newpage

% Commenting out the following part is to make loading faster

% \newpage
\appendix
\section{Parametric Study on $n$}\label{chp:n_study}
\begin{figure}[!hb]
  \centering
  \includegraphics[trim = 0in 0in 0in 0in, clip, width=\columnwidth]{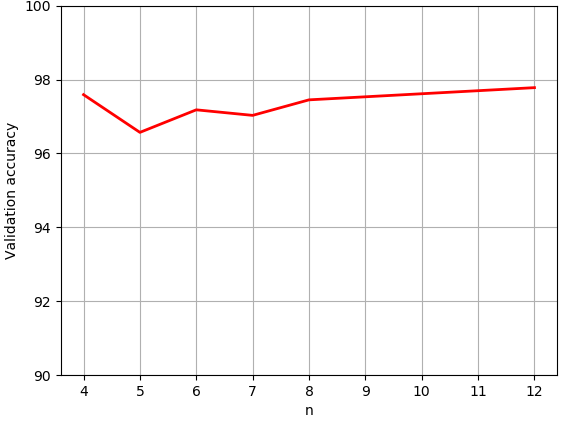}
  \caption{Validation accuracy of GS model trained on datasets with the number of sampled points $n$ increases.} 
  \label{fig:cs_n_study}
\end{figure}

In this study, we vary the number of sampled points on each vector $n\in \{4,5,6,7,8,10,12\}$ to explore its effect on the final model performance. Like GS3 model detailed in section \ref{chp:cs_res}, as $n$ increases, we enlarge the width of each layer accordingly. The resulting validation accuracies as $n$ increases are summarized in Fig. \ref{fig:cs_n_study}. From the figure, it can be concluded that there are no significant changes in validation accuracy as $n$ varies. A potential cause lies in the fact that the majority of the obtained vectors are straight lines. There is no extra useful information added to the input when more points are sampled in between. The difference when $n=4$ and $n=12$ can be ignored. But $n=12$ requires 3x more parameters to train, which usually leads to much more training time and less stability. As such, we choose $n=4$ for all of the experiments for feature exaction.

\section{More Results on 3-class Segmentation}\label{chp: more_results}
\begin{figure*}[!ht]
  \centering
  \includegraphics[trim = 0in 0in 0in 0in, clip, width=1.1\textwidth]{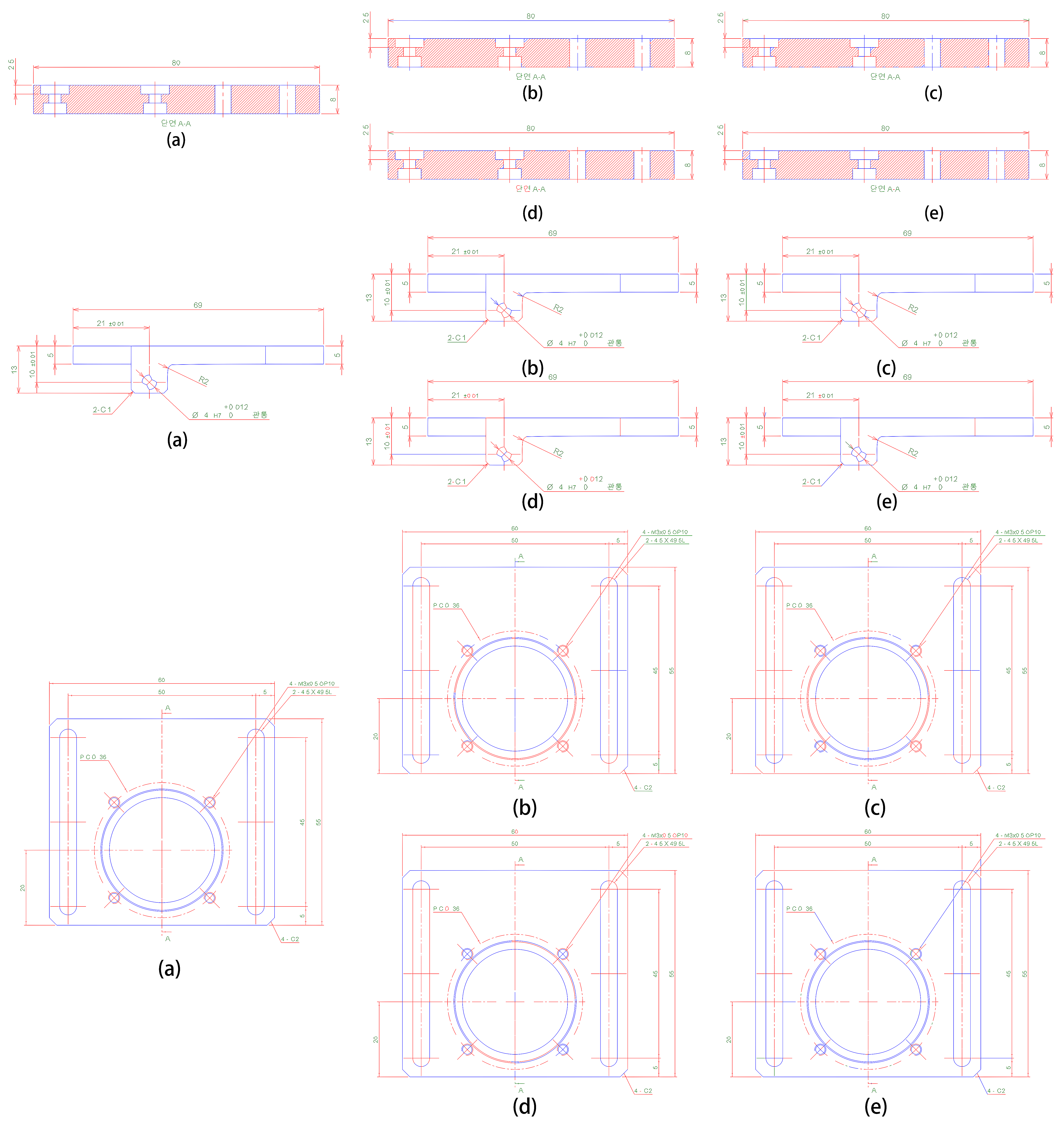}
  \caption{\txtRed{Sample prediction results from three baseline models versus ours. (a) The ground truth. (b) PSPNet results. (c) DeepLabV3 results. (d) Sketchgnn results (e) Ours. In the first part, our model yields the best results when identifying all the internal surfaces of the holes as contour lines (blue). In the second and third part, our model is the only one that correctly identifies the outline of holes as blue and the center lines on the holes as red. This is crucial when extracting the shape of the entire part.}}
  \label{fig:baseline_compare_more}
\end{figure*}

\begin{figure*}[!ht]
  \centering
  \includegraphics[trim = 0in 0in 0in 0in, clip, width=\textwidth]{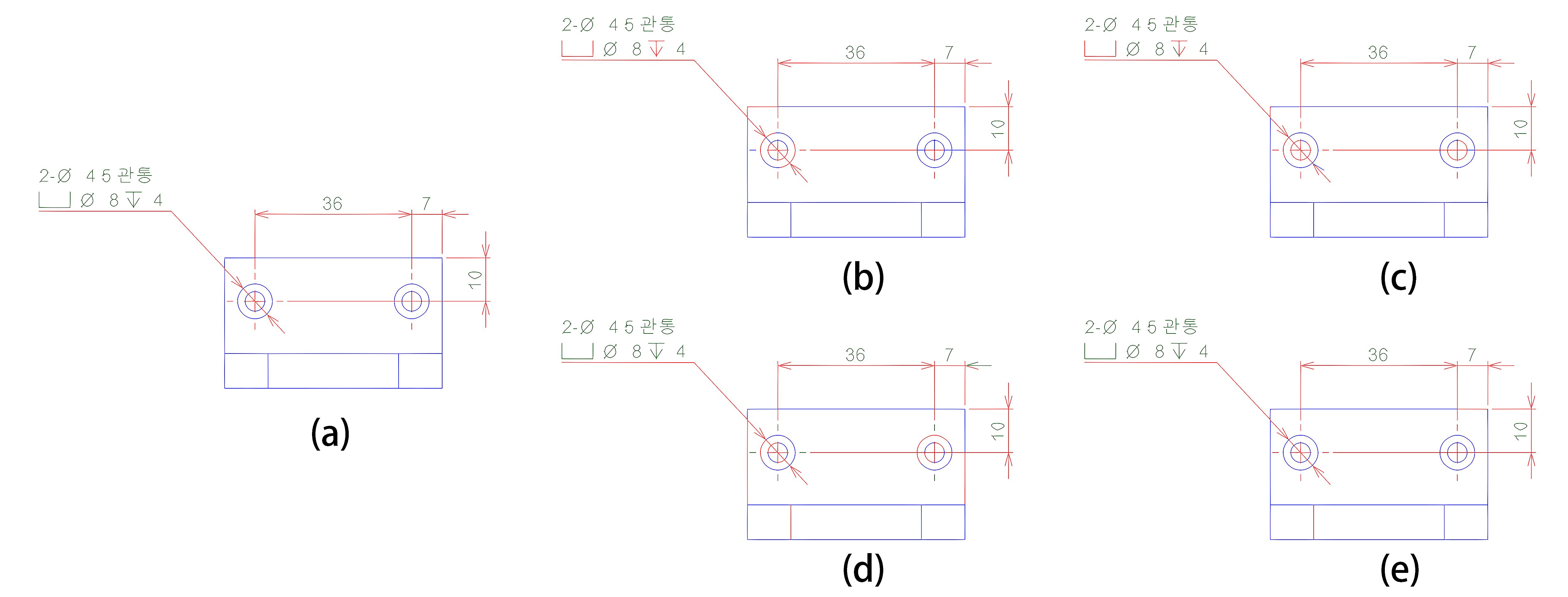}
  \caption{\txtRed{Sample prediction results from three baseline models versus ours. (a) The ground truth. (b) PSPNet results. (c) DeepLabV3 results. (d) Sketchgnn results (e) Ours. In this part drawing, our model is that only one that correctly distinguishes the outline of the holes as blue and the center lines on the holes as red.}}
  \label{fig:baseline_compare_more_ext}
\end{figure*}

As a supplement to the baseline comparison results shown in section \ref{chp:baseline}, more visual comparisons are demonstrated in Fig. \ref{fig:baseline_compare_more} and Fig. \ref{fig:baseline_compare_more_ext}.

%%%%%%%%%%%%%%%%%%%%%%%%%%%%%%%%%%%%%%%%%%%%%%%%%%%%%%%%%%%%%%%%%%%%%%%%
\newpage

\newpage

\bibliographystyle{elsarticle-harv} 
\bibliography{reference}
\end{document}

%% file: tex/1Introduction.tex
Engineering technical drawings \txtRed{of mechanical parts} serve as a universal medium for information exchange between designers and manufacturers. Such drawings encode the topological information, dimensions, and manufacturing requirements of a product in a unified and standard form, which can then be utilized in various engineering applications including content-based part indexing \citep{fonseca2005content,kasimov2015individual}, cost estimation \citep{sajadfar2015hybrid}, and process planning \citep{kulkarni2000review}. Although the underlying designs are commonly created in a vector format through digital design tools, a raster drawing is more frequently used by manufacturers due to the ease of information exchange and quality assurance. According to a survey of Japan's manufacturing industry \citep{mits2019survey}, 84\% of the customers use 2D raster-based drawings such as PDF, paper, or fax format when placing an order for manufacturing, which results in a major impediment in the automation of the aforementioned applications due to the need for human involvement in interpreting these drawings. 

\begin{figure}[!ht]
  \centering
  \includegraphics[trim = 0in 0in 0in 0in, clip, width=\columnwidth]{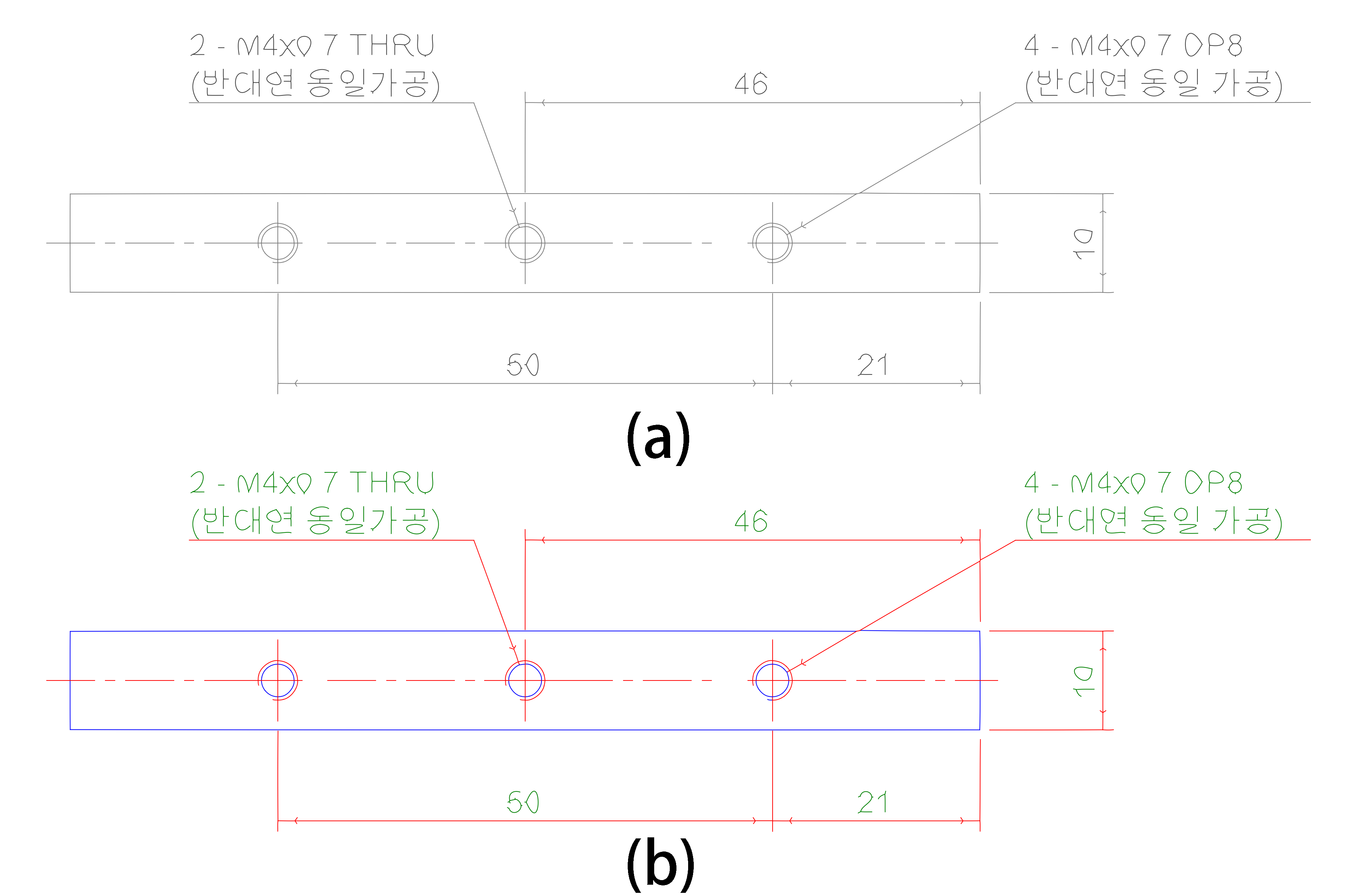}
  \caption{(a) Typical \txtRed{mechanical} engineering drawing. (b)Semantically labeled results. Blue: Contours, Green: Texts, Red: Dimension sets.} 
  \label{fig:prob_state}
\end{figure}

For a modern online platform of part manufacturing, clients often upload their designs in raster image format for better quality assurance and IP protection since the information in image drawings is noneditable. Unlike a vector format, which enables trivial digital access to all stored information through a script file, raster drawings usually require manual inspection by technicians to extract the information required for quotation and manufacturing. The inspection process includes the identification of the part shape, dimensions, and manufacturing requirements.

\begin{figure*}[!ht]
  \centering
  \includegraphics[trim = 0in 0in 0in 0in, clip, width=\textwidth]{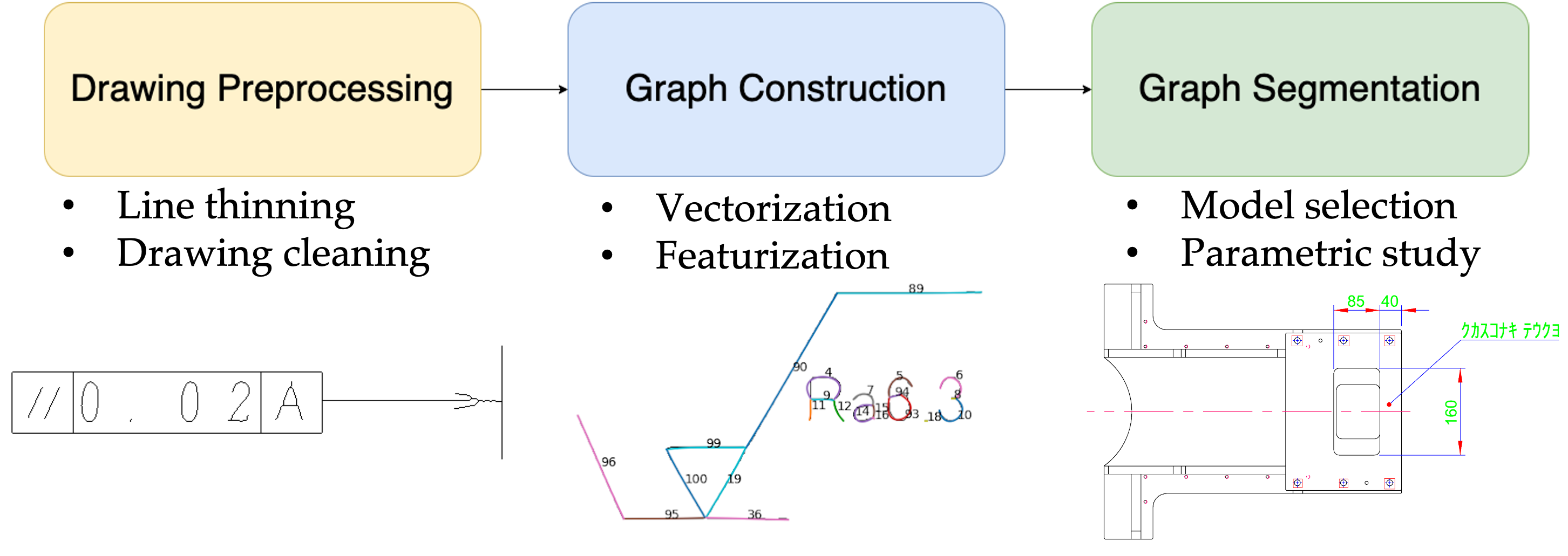}
  \caption{Our proposed workflow.}
  \label{fig:cs_pipeline}
\end{figure*}

Here, we focus on the problem of semantic segmentation of the components in raster drawings. Common \txtRed{mechanical} engineering components consist of straight lines, arcs, and circles. Our goal is to develop an automated data-driven framework that learns to distinguish between contour shapes, dimension sets, and text at the component level. Our approach improves the efficiency of the technical drawing interpretation, relieving the burden of human operators by reducing the repetitive and tedious task of labeling drawings.

While recent vision-based methods have shown to be effective in image interpretation tasks such as object detection \citep{redmon2016look,he2018mask,wang2020deep}, semantic segmentation \citep{long2015fully,ronneberger2015unet,chen2018encoderdecoder}, and visual question answering \citep{li2019visualbert,jiang2020defense,wang2021vlmo}, it is challenging to apply these methods to the segmentation of components in engineering drawings since the type of a component is also dependent on the contextual information. The same component can have different semantic meanings hence categories in different contexts. \txtRed{Mechanical} engineering drawings typically involve lines or curves only sparsely filling the image frame, and without color or textual features, making traditional image-based segmentation approaches ineffective at engineering drawing interpretation.     

To address this issue, we present a new approach to engineering drawing segmentation that maps the components in the drawing to a graph structure embedded with contextual information (Fig. \ref{fig:cs_pipeline}). The components are obtained through drawing preprocessing and a vectorization process, which then forms the graph nodes. The relationships between the components are encoded in the edges of the graph. A set of features are computed as nodal attributes to embed the shape, size, curvature, and neighboring information. With this, the task of component segmentation of a raster drawing image is formatted as a node-labeling problem.

Graph Convolutional Networks (GCNs) have shown great promise in node classification in various graph-structured data including academic networks \citep{bhattacharya2007collective}, social networks \citep{fonseca2005content}, citation networks \citep{sen2008collective} and medical data \citep{fakhraei2016adaptive, namata2012query}. Like CNNs, GCNs aggregate information from a node and its neighboring nodes using a trainable unstructured feature map, which makes it applicable to graph data of any shape. In this work, we build our data-driven model based on GraphSAGE \citep{hamilton2017inductive}, a recent GCN model that aggregates the nodal information from the neighborhood structure, to predict the component type of each vectorized entity in raster drawings. The effectiveness of our graph representation is validated by the comparison between our proposed GCN methods and three vision-based or graph-based model for image segmentation. The depth of our proposed model is optimized through a parametric study of the number of convolutional layers in the model. Results indicate superior performance in both 2-class classification and 3-class classification tasks compared to our baseline models.

%% file: tex/Contributions.tex
Our main contributions include:

\begin{itemize}
    \item[$\bullet$] A vectorization method for raster drawings by skeletonizing, tracing, splitting, and cubic bezier curve fitting.
    \item[$\bullet$] A graph representation for the extracted components in engineering drawings embedded with domain-specific nodal attributes and contextual information.
    \item[$\bullet$] A data-driven framework that takes the vectorized component graphs as input and identifies the nodal component type for semantic segmentation. 
\end{itemize}

%% file: tex/2Relatedworks.tex
In this section, we review some prior works for the analysis of engineering drawings. In addition, as inspiration to our proposed framework, we also introduce recent advances in graph-based methods for image analysis and general data-driven methods for graph node classification. 

\subsection{Content-based Methods for Engineering Drawing}
In the analysis of engineering drawings, content-based methods are broadly used for drawing match or retrieval. The key is to comprehend the basic elements in the drawings and define a measure of similarity. To detect the basic elements, Hough line transform \citep{Med2000Content}, pixel blocks \citep{jiao2009an}, and patch groups \citep{liu2010shape} are utilized as representations for extensive matching or retrieval tasks. \txtRed{Other works also focus on content-based detection for certain components like shape contour \citep{kuchuganov2020clustering}, symbols \citep{hu2021detection, elyan2020deep} and information tables \citep{sulaiman2012study} in the drawing.} When an exemplar drawing is given, the matching process seeks the closest drawing in an existing pool under a similarity measure. Prior works propose distances between a graph of decomposed topology \citep{sousa2010sketch}, distances between feature vectors \citep{Med2000Content}, weighted cosine similarity \citep{feng2009line}, and the Bhattacharya correlation between histograms \citep{huet2001relational} to identify the most relevant drawing. 

All works above are applied to drawings with only contour shapes, which restricts the extensive usage in our problem where the dimensions and texts are also included. But the idea of constructing a graph of basic elements enables a more efficient representation for raster drawings compared to the original image format.

\subsection{Data-driven Methods for Graph Node Classification}

When the graph of components is introduced, the problem of component segmentation is converted to the classification task of the graph nodes. Node classification is a critical problem in many supervised or semi-supervised learning scenarios \citep{zhu2005semi}. Various methods have been proposed for node classification including iterative classification \citep{sen2008collective}, label propagation \citep{xiaojin2002learning}, and SVM on nodal embeddings \citep{grover2016node2vec}. However, recent work has shown the promise of a better classification if the nodal embedding is jointly learned with the training of the data-driven classifier \citep{yang2016revisiting}, which drives the development of end-to-end graph neural networks. 

Since the first deep learning based framework, Deepwalk \citep{perozzi2014deepwalk}, was introduced to solve the nodal classification problem, graph neural networks have demonstrated their superior capability in efficient feature extraction on unstructured graph data. Recently, Graph Convolutional Networks (GCNs) \citep{welling2016semi} have yielded better performance due to the unique nodal information aggregation mechanism. Other variants of GCN extensively introduce various ways for aggregation including neighborhood aggregation \citep{hamilton2017inductive} and attention mechanism \citep{velivckovic2017graph}, which inspires us to build our proposed data-driven model for component segmentation.

\subsection{Graph-based Methods for Image Analysis}

For a more efficient and structured feature representation, graph models have been frequently introduced to multiple major tasks in image analysis. \cite{monti2017geometric} propose the first GNN-based method in image classification. More recent works focus on achieving image segmentation using the graph of pixels \citep{shi2000normalized} or graph of superpixels \citep{stutz2018superpixels}. These oversegmented, simpliﬁed, images can be applied to many classic tasks in computer vision, including depth estimation, segmentation, and object localization \citep{achanta2012slic}.

The biggest advantage of introducing the graph representation in image analysis is that the problem can be decomposed into a higher level image processing with a more coarse unstructured data format. Unlike the image space where each pixel is treated as an independent element, the connection between the graph nodes enforces the possible dependencies due to the continuity of images. \txtRed{\cite{paliwal2021ossr} proposed a symbol detection method based on Graph Convolutional Networks for piping diagrams. The one-shot learning model yielded comparable performance to previous fully supervised model. In a more recent work, \cite{rica2023zero} introduced a zero-error digitization approach for piping diagrams by creating a graph of symbols and components. The proposed tools were able to indicate incorrectly identified components to aid manual validation as well as search groups of components based on a given query. In both works, graph-based feature extraction methods have demonstrated their superior performance in parsing contextual information in structured diagrams. Despite the similarity of these works to our work, their primary focus is on symbol recognition, while the focus of our proposed method is on component node classification.} In mechanical engineering drawings, the type of a component is heavily dependent on the contextual information and the neighboring components. As an initial attempt to apply graph-based methods in the analysis of \txtRed{mechanical parts}, \cite{xie2022graph} developed a data-driven framework that can identify the manufacturing process of a part using a graph of detected straight-line segments in the engineering drawing. But the graph is not suitable for a complete semantic interpretation of all the components. First, only straight-line segments are used to vectorize the drawing, which leads to inaccurate vectors for circles, arcs and curved strokes in text. Second, The featurization of the vectors only contains location information (X, Y coordinates). There are no indicators to describe the topological feature (size, angle, curvature) of each obtained vector. Therefore, this work is shown to be effective in graph classification tasks (manufacturing method classification), but it is challenging to be directly extended to graph segmentation tasks (component interpretation).

As such, we propose a novel method to vectorize the drawing as a lower-level component representation with lines and curves, and construct a graph of such vectors as the basic element with topological featurization for drawing analysis, embedding the contextual information in the edges of our component graph. The proposed graph representation is utilized to achieve a component segmentation of engineering drawings.

%% file: tex/3techical_approach.tex
In this work, we present a pipeline to preprocess engineering drawings, construct a labeled drawing dataset and train a graph-based neural network model for component segmentation. This pipeline starts with a method that converts the raster drawing images into vectorized curves. Subsequently, a self-defined component graph is constructed based on the connections and distances among the obtained components. For each node (component) in the graph, we also present a novel featurization method to generate a series of feature parameters based on sampled points. Finally, this feature-embedded graph is utilized as input to a graph convolutional neural network that is able to predict the component type of each node (vector) in the drawing. Our inference pipeline is summarized in Algorithm. \ref{alg:edgnet} 

\input{tex/algorithm1}

\subsection{Drawing Vectorization}

As mentioned in Sec. \ref{intro}, vision-based methods are not effective for feature extraction from raster engineering drawings since the information exists sparsely as discrete black pixels. \txtRed{Engineering Drawings which include sheet metal parts, lathing parts, and general machining parts in their raster format do not contain any component type information and are greyscale.}
Convolutional neural networks have severe limitations in embedding distant contextual information on very large images due to their ordered grid structure \citep{simonyan2014very} and require endless increasing in depth. However, engineering drawings in their original vectorized form are capable of encoding long-scale connectivity through an arbitrary graph structure. The basic elements of this graph structure consist of lines, curves, and texts of nodes and connectivity between them. As such, we propose a novel method to vectorize the raster drawings before analysis. The method consists of three broad steps: skeletonization, trajectory tracing, and curve fitting.

\begin{figure}[!ht]
  \centering
  \includegraphics[trim = 0in 0in 0in 0in, clip, width=\columnwidth]{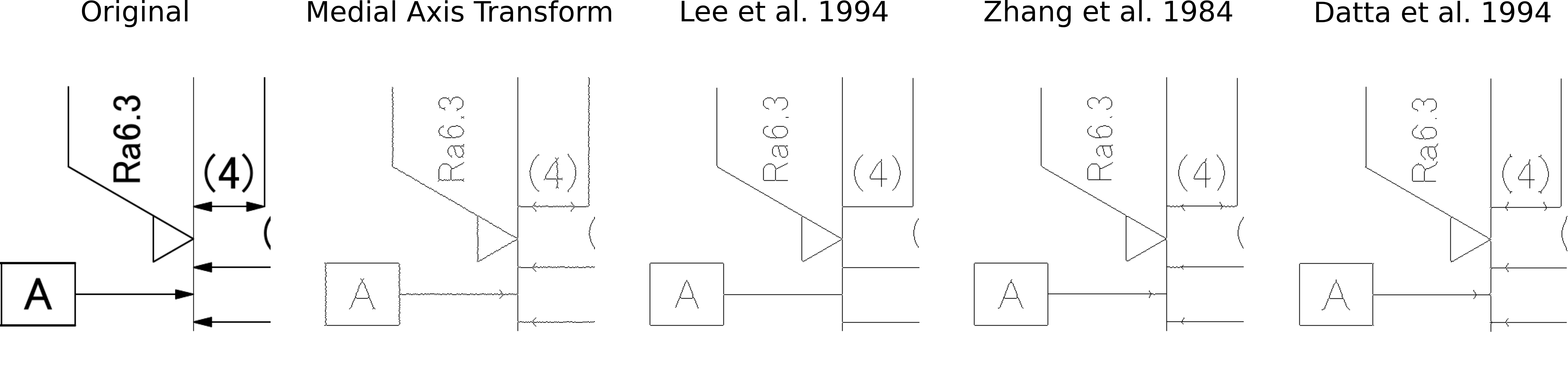}
  \caption{Comparison of four thinning methods in skeletonizing engineering drawings.}
  \label{fig:cs_skeleton}
\end{figure}

A CAD drawing is a collection of parametrically stored entities consisting of headers, blocks, tables, entities, and objects that can be easily rendered into a raster drawing through lines, curves, and text. However, it is challenging to identify these parameters from raster drawings due to the width of each entity and binary colors in the rendering process. \txtRed{It is also challenging to vectorize entities that are overlapping or lines that are intersecting, as the vectorization input only receives the rendering of the engineering drawing and not the originally created file from the CAD or drafting tool.} As such, it is necessary to morphologically thin the drawing to identify the central trace of these strokes in raster drawings and extract the skeleton of each line for better fitting. In our implementation, three line thinning algorithms \citep{zhang1984fast, lee1994building, datta1994robust} were tested for extracting the skeleton out of the original drawing (Fig. \ref{fig:cs_skeleton}). From the tests, it was concluded that the Medial Axis Transform method tends to generate more small branches where the line width is relatively large. The method from Lee's cannot retain a complete structure for small components like the arrowheads. Zhang's method does not produce clean junction points. Datta et. al. method was selected to have the most desirable thinned morphology for further processing and generating the parametric curves.

\begin{enumerate}
\item Thinning or skeletonization of the image to form traces with a single pixel width.
\item Smoothing the pixels to obtain a set of traces.
\item Splitting the traces to account for corners within the trace.
\item Removing small traces, and merging junction points
\item Fitting cubic bezier curves
\end{enumerate}

\begin{figure}[!ht]
  \centering
  \includegraphics[trim = 0in 0in 0in 0in, clip, width=\columnwidth]{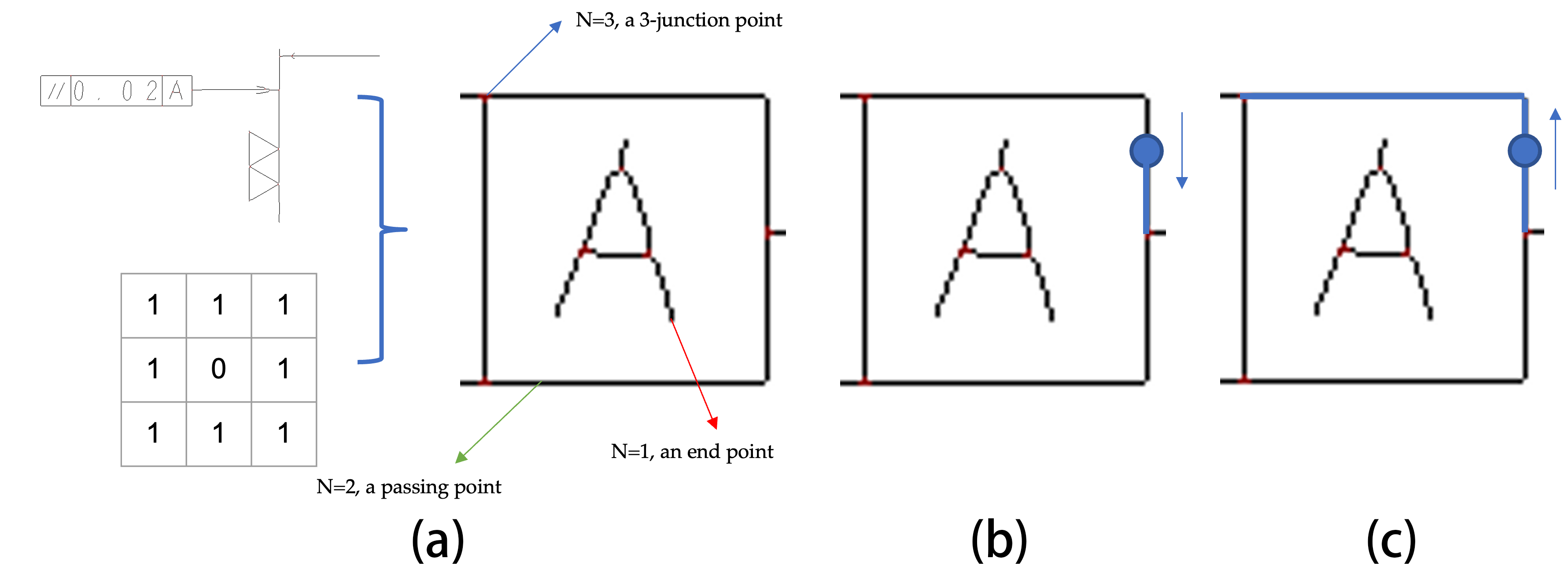}
  \caption{(a) The process of identifying the junction points, passing points and end points of a skeletonized drawing. (b) The trajectory tracing process from a passing point to a junction or an end point. (c) After (b), another tracing is done to search for the other end of the trajectory.} 
  \label{fig:cs_junction}
\end{figure}

Through the skeletonization process, strokes of lines and curves in the raster drawing are thinned to single pixel-wide trajectories. This allows us to trace the trajectories through neighboring pixels and convert the entire drawing into a unified set of parametric curves for better segmentation. In Fig. \ref{fig:cs_junction}, we summarize three types of points existing in the obtained skeleton. The point type is defined based on the number of black pixels it connects to, which is efficiently calculated by a filter scanning over all the black pixels (Fig. \ref{fig:cs_junction}(a)). 

We define a trace as an ordered list of connected pixels starting from either a junction point or endpoint and ending at either. To obtain this set of traces, we start at a randomly selected black pixel (Fig. \ref{fig:cs_junction}(b)) and iteratively visit its neighboring points until a termination point (a junction or endpoint) has been reached. We then reverse the trace to reach the other termination point. This trajectory is recorded, forming a completed trace used for parametric curve fitting. We continue this process until all black pixels have been visited and an entire set of traces have been created.

Now, with the set of pixel traces from the image, we split the traces if an edge is detected within the trace. To break the trace, we evaluated the cosine angle of the vectors formed between the point on the trace $p_i$ to the two terminal points; $p_s$ and $p_e$, given by;
$$\theta_i = \cos^{-1}(\overline{p_s-p_0}\cdot\overline{p_e-p_0}/|\overline{p_s-p_0}|\cdot|\overline{p_e-p_0}|)$$ 
Taking the second derivative of this angle provides a clear indication of a corner within the trace by forming a spike in the second derivative. We find these spikes in the $\overline{\theta}$ vector by finding the local maxima by a simple comparison of neighboring values. The trace is split at this spike point $p_i$. All the traces are split based on these criteria. 

\txtRed{Subsequently, we }check for small traces and eliminate them. Traces with $m_j<4$ pixels are eliminated to ensure that we don’t have an underdefined problem while fitting the cubic bezier curves in the next step. We also merge the terminal points of small traces, so the neighboring traces are now connected. We also merge junction points to ensure the edge-connectivity of the graph. Finally, we fit cubic bezier curves on traces. We use the least square method for fitting the cubic Bezier curves as shown in the equation:
$$\overline{\mathbf{p}}^j =[p_{0}^j, p_{1}^j, \cdots, p_{m_j-1}^j]^T$$
$$B_{i,n_{ord}}^j(t) ={n_{ord} \choose i}(t)^i(1-t)^{n_{ord}-i}, i=0,1,\cdots n_{ord}, t\in[0,1]$$
$$\mathbf{B_{m_j\times(n_{ord}+1)}^j}\mathbf{x_{(n_{ord}+1)\times2}^j}=\mathbf{\overline{p}_{m_j\times2}}$$
where, $\mathbf{\overline{p}}^j$ is the ordered list of $m_j$ points in the $j^{th}$ trace, $B_{i,n_{ord}}^j(t)$ is the Bernstein polynomial of order $n_{ord}=3$, $\mathbf{B^j}$ is the Bernstein matrix of the cubic bezier curve and $\mathbf{x^j}$ is the control points for the $j^{th}$ trace.
We set the terminal points as the first and last points of the bezier curve control points.i.e. $x_{0}^j=p_{s}^j$ and $x_{3}^j=p_e^j$.

\begin{figure}[!ht]
  \centering
  \includegraphics[trim = 0in 0in 0in 0in, clip, width=\columnwidth]{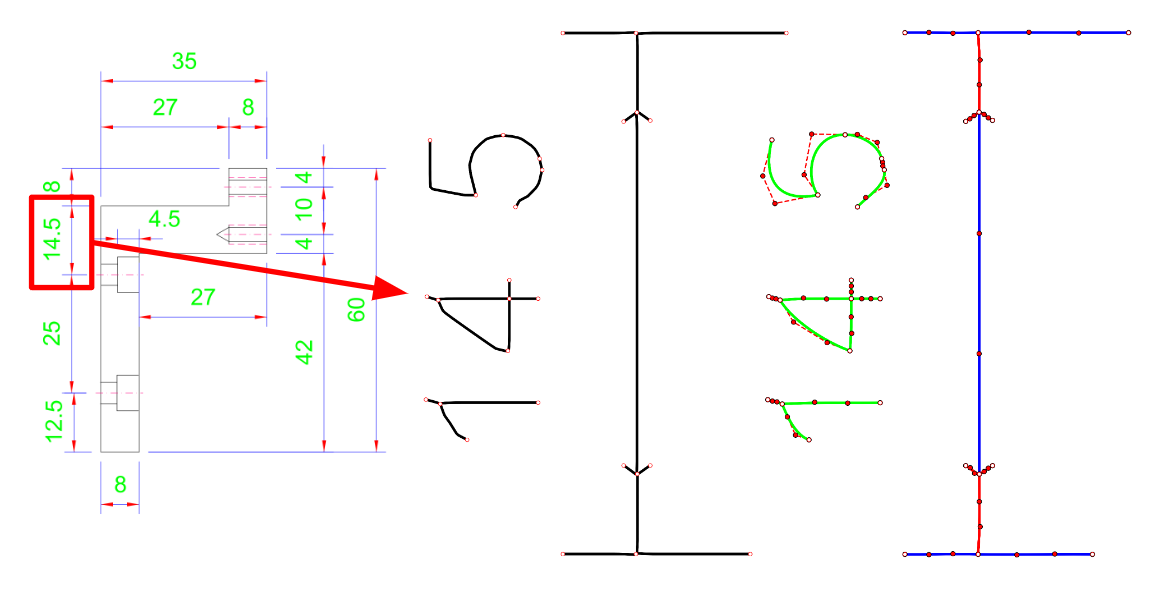}
  \caption{Raster drawing with the highlighted region showing the thinned pixels with junction points and subsequently the cubic bezier curves fitted on the traces with their respective control points.} 
  \label{fig:vecBezier}
\end{figure}

\begin{figure}[!ht]
  \centering
  \includegraphics[trim = 0in 0in 0in 0in, clip, width=\columnwidth]{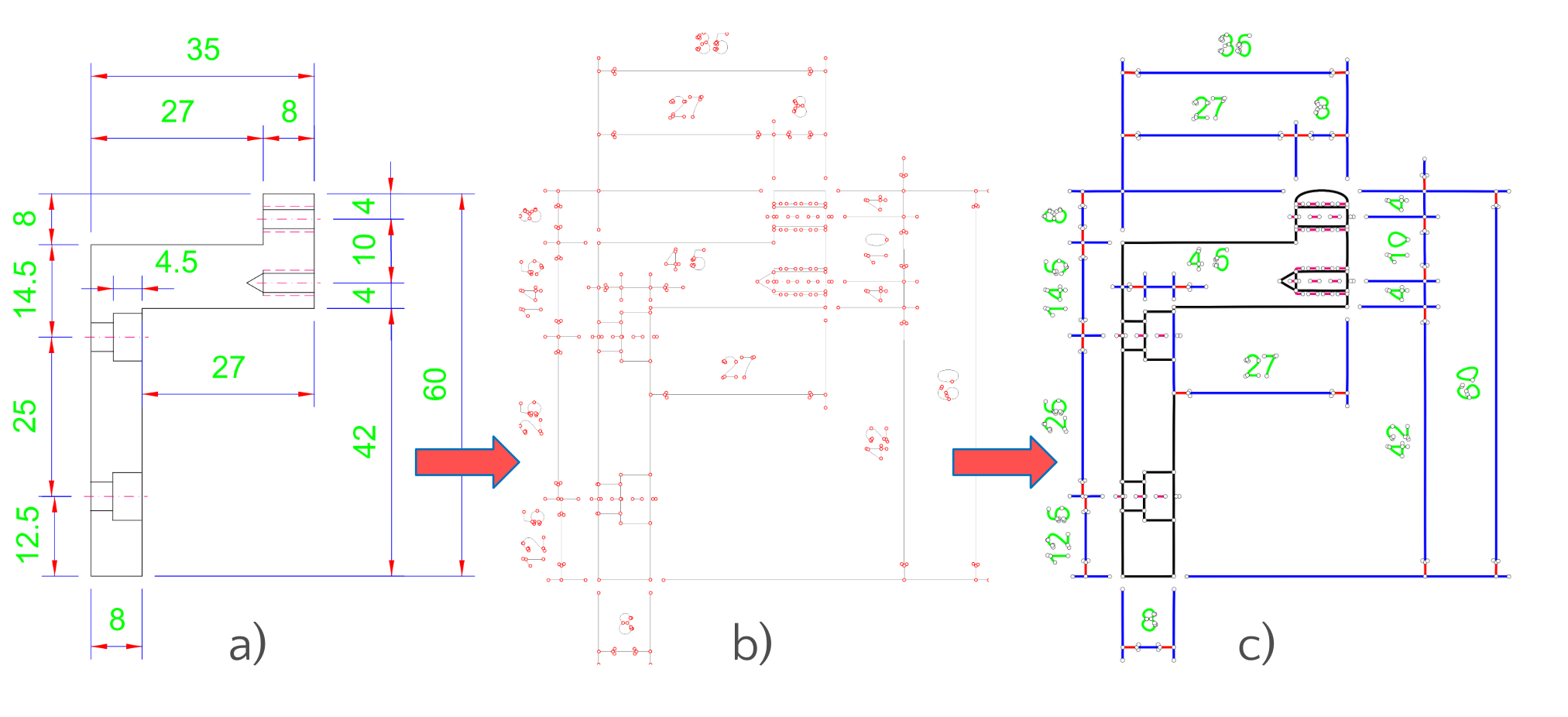}
  \caption{(a) Raster drawing (b) Thinned, smoothed, and traced trajectories with junction points marked in red circles (c) Cubic bezier curves fitted on pixel traces.} 
  \label{fig:vecSteps}
\end{figure}

These steps provide the necessary set of parameterized vectors needed for the construction of a unified graph used in our approach for segmenting the engineering drawing into its various components.

\subsection{Graph Construction} \label{chp:graph construction}

At the heart of our proposed method, a graph representation for the components is the key to embedding topological features and contextual information among the distant but connected components. In our graph structure, the node vectors are obtained from our vectorization. The graph edges are generated from the connection between these vectors. To form a unified representation, we sample $n$ evenly spaced points along each vector for featurization. Tab. \ref{tab: node_feature} lists the features computed based on these sampled points. The proposed features encode the shape, length, angle, and curvature information from each vector. Note that the entire drawing is normalized to fit in a unit square before being fed into the featurization process to ensure that all features listed are independent of the input drawing size. In summary, our graph model is defined as:
$$
G(N,E), N\in \mathbb{R}^{n_N\times(5n-1)}, E\in\mathbb{Z}^{n_E\times2}
$$
where, $N$ are the nodal features of a graph with $n_N$ nodes, $E$ are the edge indices of a graph with $n_E$ edges. In this way, the target task of this work is converted to predicting the nodal labels. The label, as well as the component type, is defined as $Y\in \mathbb{Z}^n_N$. It is a categorical parameter list indicating the type (contour lines, dimension lines, or texts)  of each vector.

\begin{table}[!ht]
\centering
\caption{\label{tab: node_feature} Our proposed nodal features. $n$ is the number of sampled points on each vector. }
\begin{tabular}{|c|p{6cm}|c|}
\hline
Feature Type            & Parameter                                                      & Dimension \\ \hline
Shape                   & $n$ points sampled along the trace, XY coordinate                & $2n$        \\ \hline
\multirow{3}{*}{Length} & Length between each pair of consecutive points                 & $n-1$       \\ \cline{2-3} 
                        & Total length (of the curve)                                    & $1$         \\ \cline{2-3} 
                        & first-to-last/total                                            & $1$         \\ \hline
Angle                   & Cos angle between each pair of consecutive short line segments & $n-2$       \\ \hline
Curvature               & Curvature at each sampled point                                & $n$         \\ \hline
\end{tabular}
\end{table}

\begin{figure}[!ht]
  \centering
  \includegraphics[trim = 0in 0in 0in 0in, clip, width=\columnwidth]{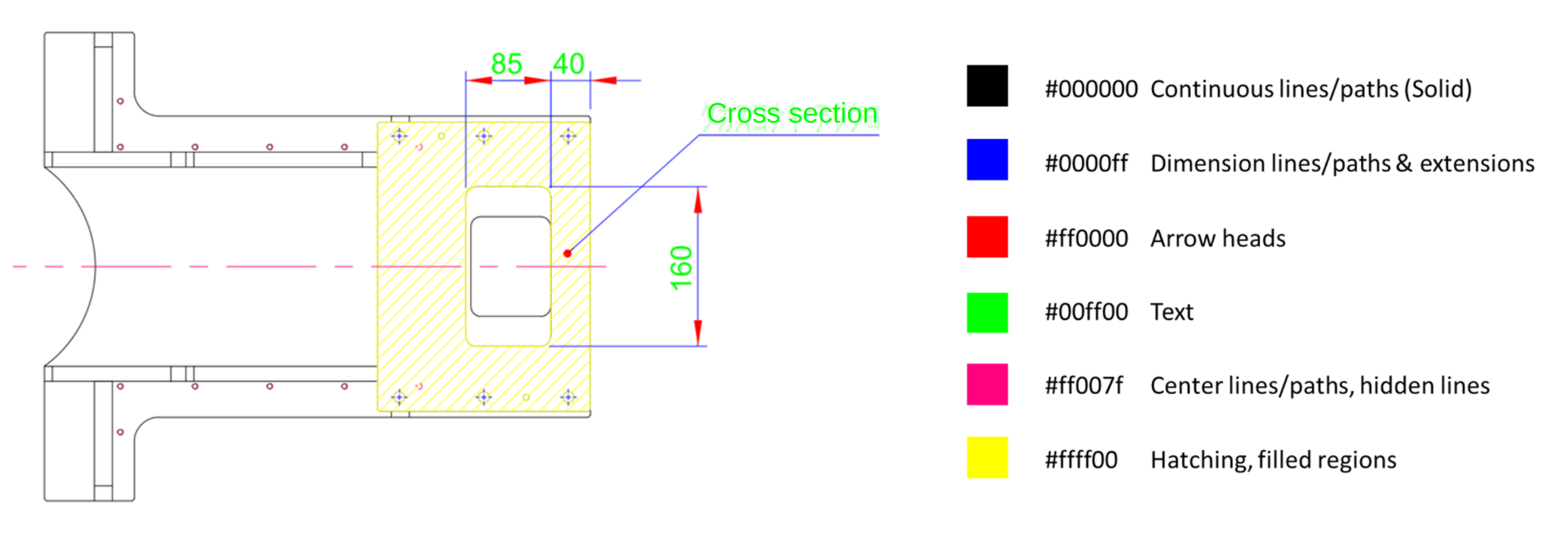}
  \caption{A sample rendering of a DXF drawing used for ground truth label retrieval. Components of each type are painted with a unique color.} 
  \label{fig:cs_gt}
\end{figure}

To establish the ground truth labels $Y_{gt}$ when creating a dataset for training, we modify a batch renderer for DXF drawings. The renderer EZDXF (\url{https://ezdxf.mozman.at/}) is modified to be able to parse the component type information stored in the vector DXF drawing and paints each component type with a unique color (Fig. \ref{fig:cs_gt}). For each vector in the graph, we sample points and check the corresponding color of each point location in the ground truth image. Finally, the ground truth label for each vector is determined by majority voting of the sampled points. In this work, we conduct two testing conditions in terms of the segmentation labels: (1) text/non-text: The model is designed to distinguish the green components vs the other ones. (2) text/contour/dimension: The model is designed to distinguish the green, the black, and the other components. The labels $Y_{gt}$ are converted to one-hot encoding accordingly.

\subsection{Graph Segmentation}

\begin{figure}[!ht]
  \centering
  \includegraphics[trim = 0in 0in 0in 0in, clip, width=\columnwidth]{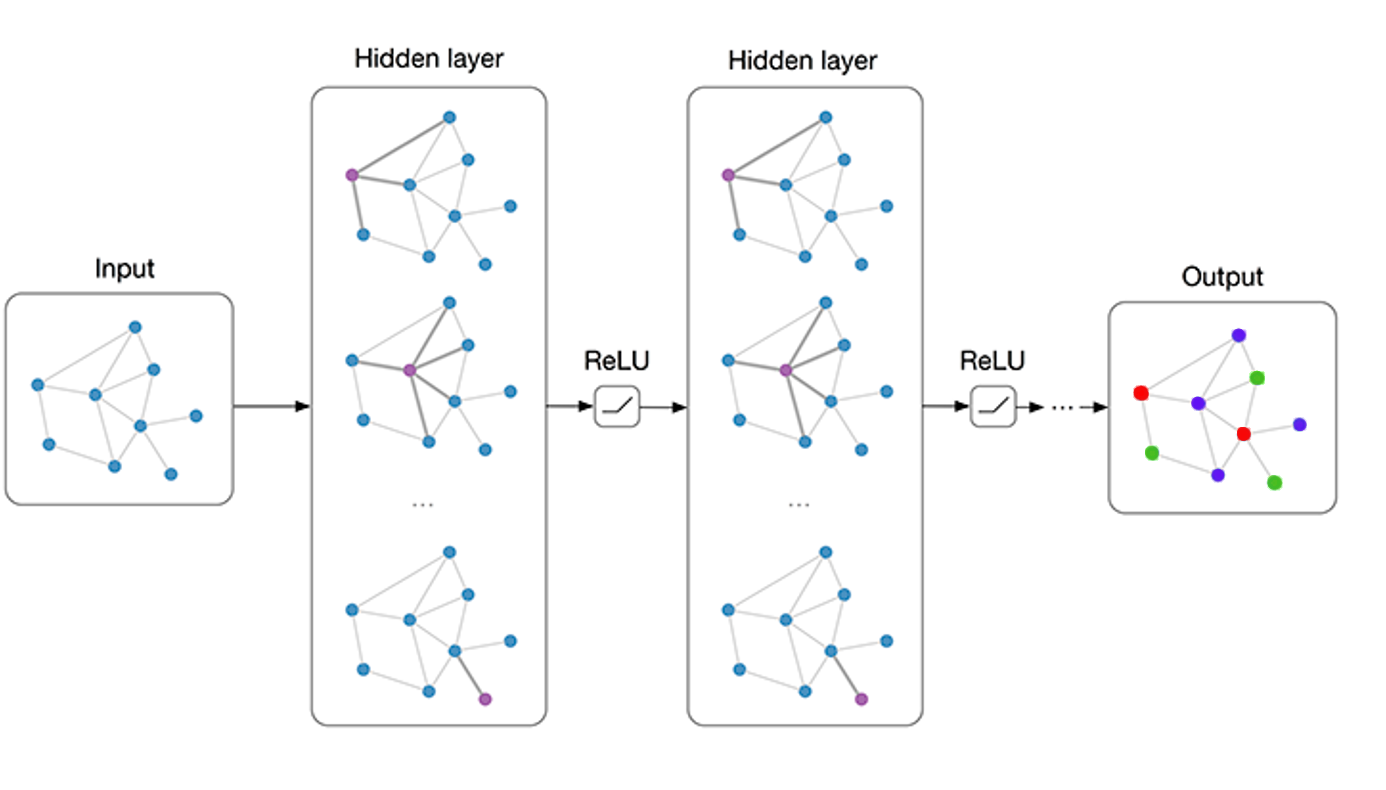}
  \caption{A general structure of \txtRed{Graph Convolutional Network (GCN) in nodal classification tasks.}} 
  \label{fig:cs_gcn}
\end{figure}

Based on the graph representation explained above, our task can be represented as:
$$
Y\leftarrow F_\theta(G(N,E))
$$
where $G(N,E)$ is our proposed graph structure, $F_\theta$ is a data-driven model with trainable parameters $\theta$, $Y$ is the predicted component type label from the model. During the training of $F_\theta(\cdot)$, $Y_{gt}$ is utilized for loss calculation. Then, we introduce a model $F_\theta(\cdot)$, a loss function $L$, and an optimizer to launch the training process.

Graph Convolutional Networks (GCNs) (Fig. \ref{fig:cs_gcn}(a)) is a type of network that works on arbitrarily structured graphs. They have been shown to be effective in various nodal classification tasks including citation and social networks \citep{welling2016semi, hamilton2017inductive}. GraphSAGE model shows superior performance in predicting the nodal class label over the other baseline graph models \citep{hamilton2017inductive}. Therefore, we build our model based on GraphSAGE. A GCN \citep{welling2016semi} model and an MLP model are also implemented to serve as baseline models. As a parametric study, we vary the number of convolutional layers in the experiments for the optimized network architecture.

Based on the steps explained above, a graph dataset is constructed based on $430$ real engineering part drawings\txtRed{, including sheet metal parts, lathing parts, and general machining parts,} \txtRed{ from a large e-commerce system for custom mechanical parts. The drawings are in DXF format originally and converted to black-and-white raster images as training data. The number of vectorized components in each drawing ranges from 500 to 3,000 }. The dataset is split into $80/20$ for training and validation. During training, the loss function is chosen as the cross-entropy loss between the predicted nodal class and the ground truth label. Adam \citep{kingma2014adam} optimizer is utilized with a learning rate $1e-3$, weight decay $5e-4$. All the models are trained with a maximum epoch of $10,000$ and batch size of $16$. The best model regarding the validation accuracy is saved for inference. In our experiments, each model takes about $20$ hours to train on a GeForce RTX 2080 TI Graphics Card.

%% file: tex/algorithm1.tex
\begin{algorithm}
\caption{Engineering Drawing Graph Network (EDGNet)}\label{alg:edgnet}
\SetAlgoLined
\SetKwInOut{Input}{Input}\SetKwInOut{Output}{Output} \SetKwData{Parameters}{Parameters}
\Input{ A raster engineering drawing $D_r$}
\Output{ A graph of vectorized components with semantic labeling $G(N,E), Y$}
% \KwData{ $N_{dg}$, the total number of generated dimension sets. $N_{do}$, the total number of dimension sets in the original drawing. $Pool_{KP}$, the pool of key points read from the original DXF file. $K^i_1, K^i_2$, two sampled key points from the pool. $O^i$, the orientation of the dimension set.  $P_b^i$, the base point for measure.  $d_{max}$, the maximum sample distance, determined by the key points location and page size. $h_f$, the font size. $H_i$, the horizontal distance between $K^i_1, K^i_2$. $B_b^m$, the bounding box of any dimension set in the hashset $S_b$.$ C_{contour}$, the bounding box of the contour shape.}

$D_s\gets$ skeletonize($D_r$)+smoothing\;
Split $D_s$ to strokes $\{S_i\}_{n_N}$ from junction points\;
Fit cubic bezier curves $\{B_i\}_{n_N}$ to each stroke $S_i$\;

Init $i=1$\;
\While{$i\leq n_N$}{
Sample $n$ equally spaced points $P_i$ on $B_i$\;
Calculate nodal features $N\in\mathbb{R}^{n_N\times(4n-1)}$\;
$i=i+1$\;
}
Assemble a component graph $G(N,E)$, $E$ are the edges (connections) between the nodes\;
Get predicted semantic labels from a graph convolution network $Y\in \mathbb{R}^n_N\gets F_\theta(G(N,E))$ 

% \While{$i\leq N_{dg}$}{
% %\Comment{iteratively generate a new dimension set}

% $K^i_1, K^i_2 \gets$ Sample($Pool_{KP}$) \;
% % \Comment{randomly pick two key points from the pool} 
% $O^i\gets$ \textbf{random}([Vertical, Horizontal])\;
% calculate distances for two placements: $d_{p1} = d(K^i_1,  C_{contour}),$ $d_{p2} = d( K^i_2, C_{contour})$\;
% Select the near side $argmin(d_{p1},d_{p2})$.\;
% $d_b^i\gets$ \textbf{random(}$[0, d_{max}]$),  $P_b^i$ is determined given the key points, orientation and direction. The bounding box $B_b^i$ of text measure for $P_b^i$ is recorded as $[P_b^i, h_f, H_i]$\;
% \If {$B_b^i \cap B_b^m \neq \emptyset$}{go to 7.\;}
% \If {$P_b^i \in C_{contour}$}{go to 7.\;}
% Save $K^i_1, K^i_2, P_b^i$, orientation(binary) to the DXF file. Save $B_b^i$ to the hashset $S_b$.\;
% }

\end{algorithm}

%% file: tex/4results.tex
This section demonstrates a series of experiments to validate the effectiveness of our graph representation. Additionally, Experiment II focuses on optimizing the model architecture with respect to the validation accuracy. Finally, this model setup is extended to a 3-class segmentation problem. 

\subsection{Model Selection}

Using our graph representation described in Section \ref{chp:graph construction}, a classifier is trained to predict the component type of each vector in the drawing. Here, we introduce three data-driven classifiers, including our proposed GraphSAGE model (GS), a vanilla GCN model (GCN) as a graph method baseline, and a Multi-layer Perceptron model (MLP) as a non-graph method baseline in our designed \textbf{Experiment I}. To maintain a fair comparison, all three models are designed to have similar architecture and depth. The details are: 
\begin{itemize}
    \item \textbf{GCN} model: 3 vanilla graph convolutional layers+2 linear layers, number of nodes in each hidden layer: $[32,64,128,32]$. $4\times$ReLU +$1\times$Softmax as nonlinear activation.
    \item \textbf{GS} model: 3 GraphSAGE convolutional layers+2 linear layers, number of nodes in each hidden layer: $[32,64,128,32]$. $4\times$ReLU +$1\times$Softmax as nonlinear activation.
    \item \textbf{MLP} model: 5 linear layers, number of nodes in each hidden layer: $[32,64,128,32]$. $4\times$ReLU +$1\times$Softmax as nonlinear activation.
\end{itemize}

\begin{figure}[!ht]
  \centering
  \includegraphics[trim = 0in 0in 0in 0in, clip, width=\columnwidth]{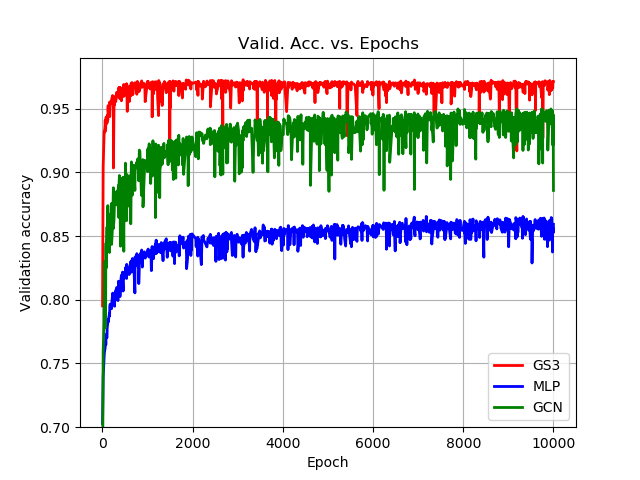}
  \caption{The training curves for GS, GCN and MLP model in \textbf{ Experiment I}.} 
  \label{fig:cs_exp1}
\end{figure}

In \textbf{Experiment I}, the three models described above are trained with identical training conditions to distinguish the text vs non-text components in our constructed part drawing dataset. We use $n=4$ for the dataset creation in this experiment based on a parametric study detailed in section \ref{chp:n_study}. Fig. \ref{fig:cs_exp1} illustrates the validation accuracy in the training process. It can be concluded that graph-based models (GS and GCN) yield better ($>5\%$) results than the non-graph-based model (MLP), which speaks for the necessity of the contextual information embedded in our designed graph structure. Conversely, GS and GCN model share the same pattern in the validation curve in the early phase of the training ($<500$ epochs). Then, the GS model rapidly converges at around 1000 epochs to $95\%$, while the GCN model gradually converges to a lower value at around 3000 epochs. The results confirm the superior capability of our implemented GS model. Next, we want to boost its performance by optimizing the model depth.

\subsection{Model Depth Study}
Through \textbf{Experiment I}, the GS model with $3$ convolutional layers yields the best performance over the other two models. Therefore, we further design \textbf{Experiment II} to study the effect of model depth on the final classification accuracy. Three GS models in this experiment are implemented with $3, 4,$ and $5$ graph convolutional layers. For consistency, all three networks are assembled so that the graph layers expand the dimensions. Then the linear layers squeeze the dimensions to the final desired number of classes. The details are:

\begin{itemize}
    \item \textbf{GS3} model: 3 GraphSAGE convolutional layers+2 linear layers, number of nodes in each hidden layer: $[32,64,128,32]$. $4\times$ReLU +$1\times$Softmax as nonlinear activation.
    \item \textbf{GS4} model: 4 GraphSAGE convolutional layers+3 linear layers, number of nodes in each hidden layer: $[32,64,128,256,128,32]$. $6\times$ReLU +$1\times$Softmax as nonlinear activation.
   \item \textbf{GS5} model: 5 GraphSAGE convolutional layers+4 linear layers, number of nodes in each hidden layer: $[32,64,128,256,512,256,128,32]$. $8\times$ReLU +$1\times$Softmax as nonlinear activation.
\end{itemize}

\begin{figure}[!ht]
  \centering
  \includegraphics[trim = 0in 0in 0in 0in, clip, width=\columnwidth]{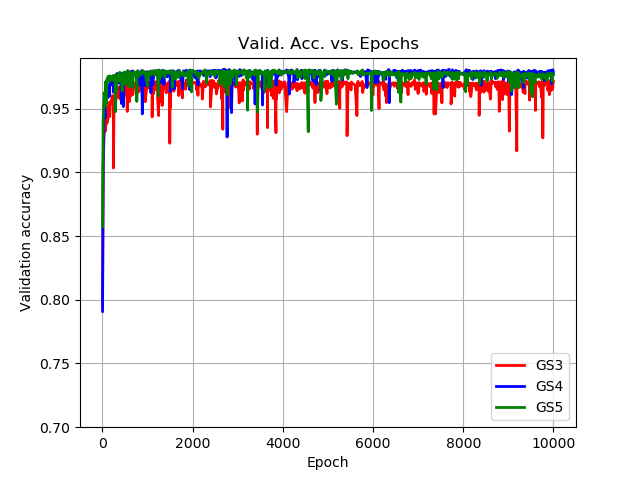}
  \caption{The training curves for GS models of different depths in experiment II.} 
  \label{fig:cs_exp2}
\end{figure}

\begin{figure}[!ht]
  \centering
  \includegraphics[trim = 0in 0in 0in 0in, clip, width=\columnwidth]{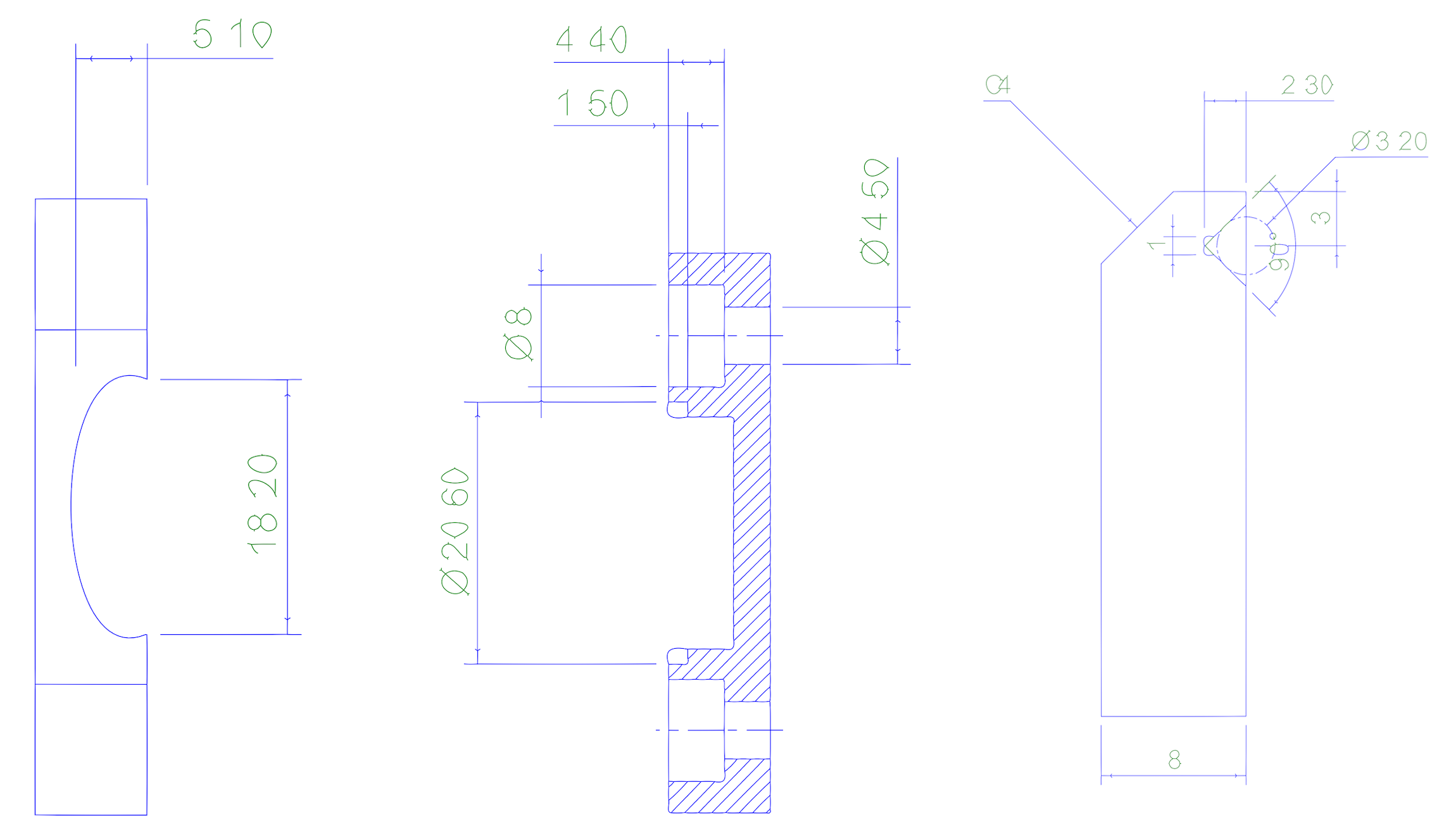}
  \caption{Sample prediction results from GS5 models in experiment II. The colors \txtRed{green} and blue indicate the predicted text components and non-text components respectively.} 
  \label{fig:cs_2class}
\end{figure}

We utilize the same setup as Experiment I for training all three models above. Fig. \ref{fig:cs_exp2} demonstrates the validation curves for these GS models during the training session. We conclude that the performance of the GS model tends to increase for deeper models. However, the improvement gradually levels out when 5 convolutional layers are used. Additionally, Tab. \ref{tab: 2class} summarizes the confusion matrix for the best model, GS5. Sample prediction results are also demonstrated in Fig. \ref{fig:cs_2class}. Results show that our GS5 model achieves nearly perfect results in all three test drawings. More statistical comparison results with other baseline models are detailed later in section \ref{chp:baseline}. Then, we extend the test condition to multi-class segmentation. 

\begin{table}[!ht]
\centering
\caption{\label{tab: 2class} Confusion matrix of our proposed GS5 model in the 2-class segmentation task. From these statistics, the evaluation results can be calculated: \textbf{Precision$=98.38\%$, Recall$=98.57\%$, Accuracy$=98.48\%$}. }
\begin{tabular}{c|cc|}
\cline{2-3}
                               & \multicolumn{2}{c|}{Prediction}       \\ \hline
\multicolumn{1}{|c|}{GT}       & \multicolumn{1}{c|}{Text}  & Non-text \\ \hline
\multicolumn{1}{|c|}{Text}     & \multicolumn{1}{c|}{21103} & 304      \\ \hline
\multicolumn{1}{|c|}{Non-text} & \multicolumn{1}{c|}{345}   & 13559    \\ \hline
\end{tabular}
\end{table}

\subsection{Multi-class Segmentation}

In previous experiments, the classifier is trained to distinguish between the text and non-text components. This task is relatively simple since text components are usually unique in terms of size and curvature compared to all other straight lines and curves. For practical use, the human inspector needs to comprehend the overall shape of the part through all contour lines, and then gather all manufacturing requirements through dimensions and texts. As such, we construct a dataset with three component types for prediction: Contour, Text, and Dimension, which correspond to black, green, and all other colored lines in Fig. \ref{fig:cs_gt}. The dataset is utilized in \textbf{Experiment III} for training a GS5 model in a 3-class segmentation task for the part drawings.

\begin{table}[!ht]
\centering
\caption{\label{tab: 3class} Confusion matrix of our proposed GS5 model in the 3-class segmentation task. Based on the statistics, the performance of the model can be calculated: \textbf{Precision: $83.03\%,95.04\%,90.70\%$, Recall: $83.46\%,96.07\%,89.60\%$, Accuracy: $90.82\%$}.}
\begin{tabular}{c|ccc|}
\cline{2-4}
                                & \multicolumn{3}{c|}{Prediction}                                       \\ \hline
\multicolumn{1}{|c|}{GT}        & \multicolumn{1}{c|}{Contour} & \multicolumn{1}{c|}{Text}  & Dimension \\ \hline
\multicolumn{1}{|c|}{Contour}   & \multicolumn{1}{c|}{4238}    & \multicolumn{1}{c|}{130}   & 710      \\ \hline
\multicolumn{1}{|c|}{Text}      & \multicolumn{1}{c|}{105}      & \multicolumn{1}{c|}{9229} & 273       \\ \hline
\multicolumn{1}{|c|}{Dimension} & \multicolumn{1}{c|}{761}     & \multicolumn{1}{c|}{352}   & 9589     \\ \hline
\end{tabular}
\end{table}

\begin{figure}[!ht]
  \centering
  \includegraphics[trim = 0in 0in 0in 0in, clip, width=\columnwidth]{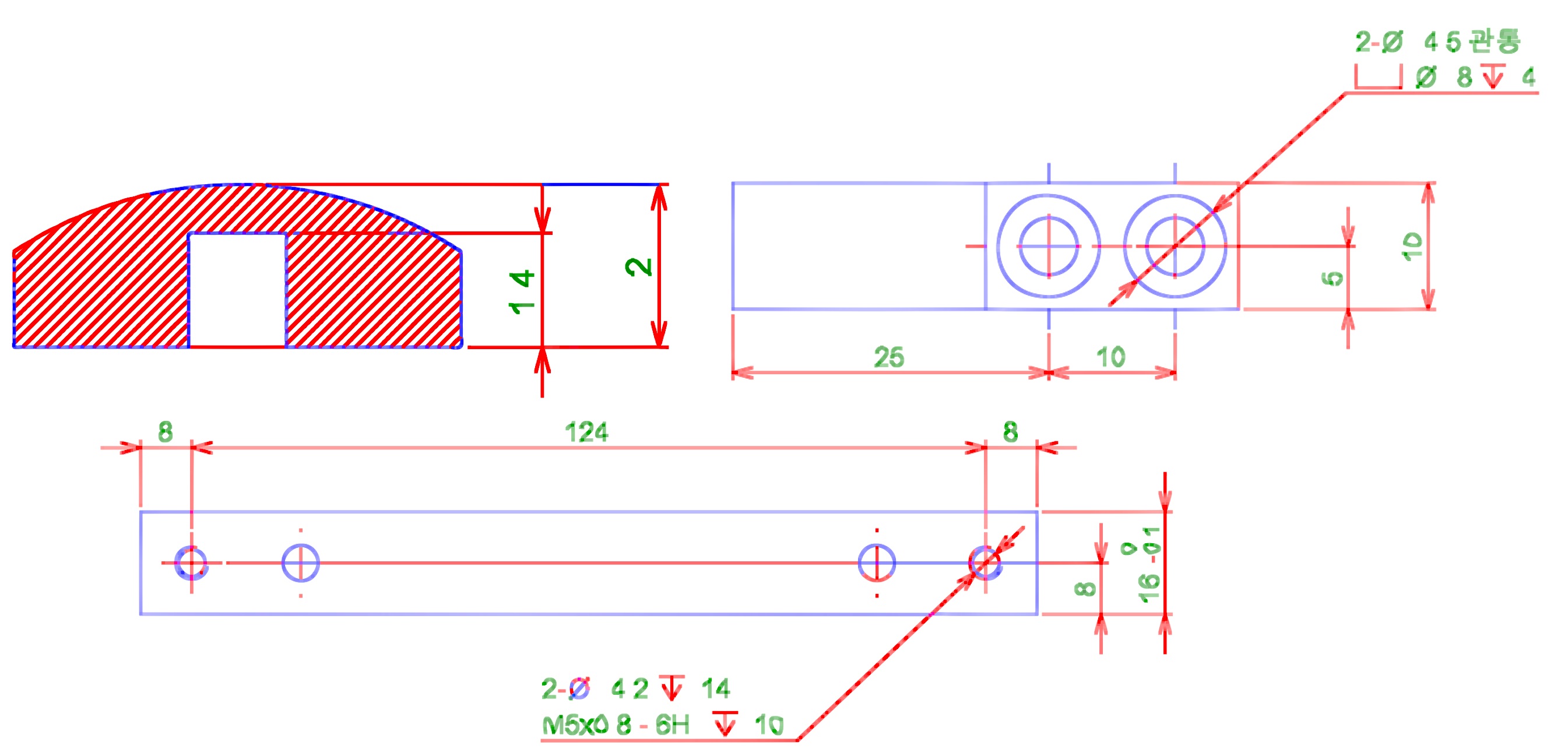}
  \caption{Sample prediction results from GS5 in \textbf{ Experiment III}. Blue, green and red indicates the predicted contour components, text components, and dimension components respectively.} 
  \label{fig:cs_3class}
\end{figure}

The resulting confusion matrix of our proposed model (GS5) is shown in Tab. \ref{tab: 3class}. It can be concluded that the model retains high accuracy on the text components, while the separation between the contour lines and dimension sets is more challenging to learn. Fig. \ref{fig:cs_3class} illustrates some failure cases in the prediction results. Two major types of failure cases are: (1) Isolated small components in the text. For example, misclassification happens on the dashed line, diameter symbol, and through hole symbols in the sample results. A potential cause is that these components are not connected with any other components in the drawing, which makes the prediction equivalent to judging the component type only by its topological features without contextual information. The issue can be resolved if we also generate edges for the nearest neighbors of each component. (2) The region where different types meet, like the contour line with its correlated extension line. The issue is likely a result of the lack of information on the connections between components. There is no indication of the difference between a line-line connection and a line-curve connection. In our graph design, a series of edge features should also be added to provide such insights into the graph network model.  

\subsection{Baseline Comparison}\label{chp:baseline}
To better understand the challenge, we construct three baseline models to achieve the component segmentation tasks with the same dataset, including two image-based deep learning models based on PSPNet \citep{zhao2017pyramid} and DeepLabV3 \citep{chen2017rethinking}, and one graph-based method based on Sketchgnn \citep{yang2021sketchgnn}. For image-based models, a black-and-white drawing image is fed to the model to predict a map of indices to indicate the semantic label of each pixel. To transfer the pixel level prediction to the component level, we map the predicted labels to vectorized results using majority voting from all pixels each vector is passing.

\begin{figure*}[!ht]
  \centering
  \includegraphics[trim = 0in 0in 0in 0in, clip, width=\textwidth]{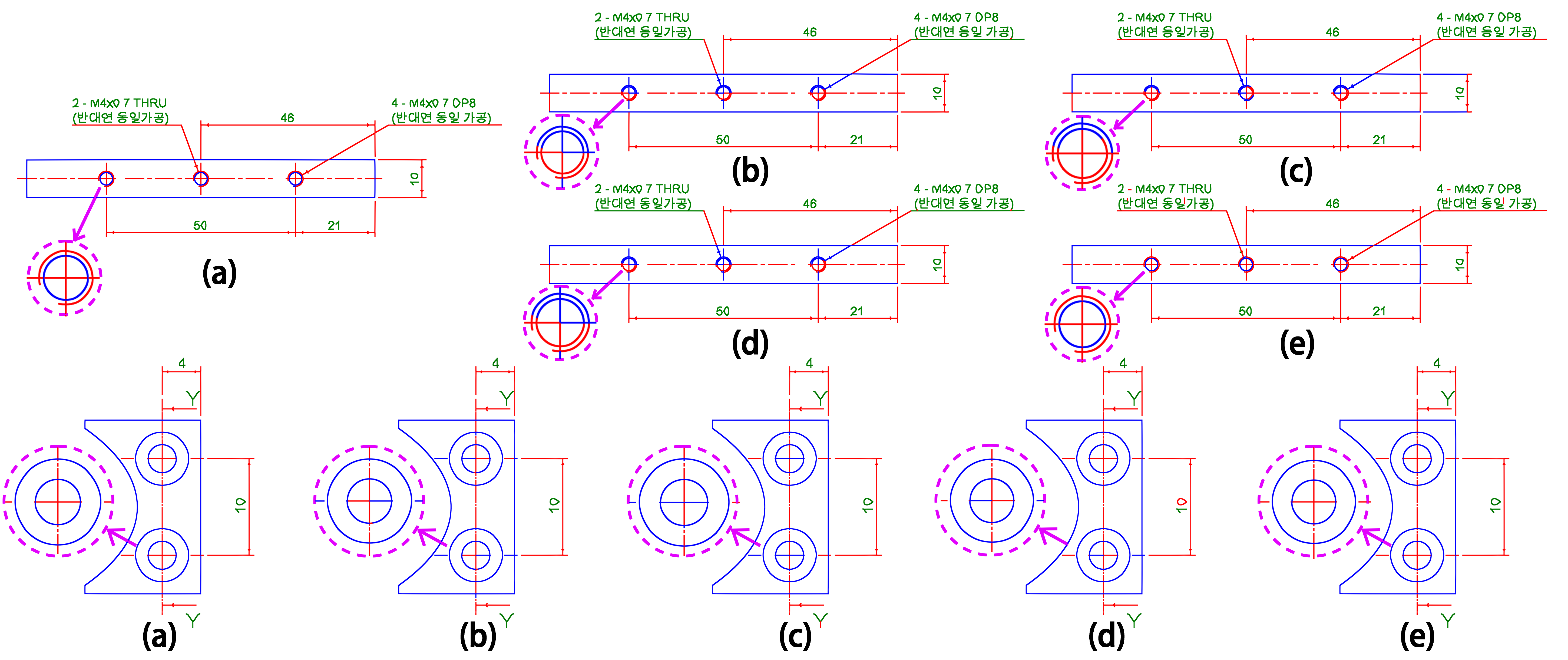}
  \caption{Sample prediction results from three baseline models versus ours. (a) The ground truth. (b) PSPNet results. (c) DeepLabV3 results. (d) Sketchgnn results (e) Ours.} 
  \label{fig:cs_compare_combo}
\end{figure*}

Using the same set of data for training and validation, the resulting validation accuracy for all three baseline models and our model are summarized in Tab. \ref{tab: acc_compare}. Here, we compare the models in three segmentation tasks: (1) Text vs. Non-text (Contour+Dimension). (2) Contour vs. Non-contour (Text+Dimension). (3)Text vs. Contour vs. Dimension. It can be concluded that our model yields the best performance in all three tasks. Evident improvement can be seen in the  separation between contour and dimension (tasks 2 and 3) when using graph-based models, which reinforces the idea that it is more challenging for image-based approaches to parsing sparse, man-made images. From the visual comparison shown in Fig. \ref{fig:cs_compare_combo}, it can be seen that our model is the only one that successfully identifies both the hole and thread line in the first example. Additionally, the other models usually have difficulty when there are multiple concentric circles with center lines. These misclassified components can easily mislead the model when extracting the overall shape of the part. More comparison results are demonstrated in \ref{chp: more_results}.

% \usepackage{multirow}
\begin{table*}[!ht]
\centering
\caption{\label{tab: acc_compare} Comparison results for \txtRed{accuracy of the predicted component labels on the validation set (\%)} in three segmentation tasks. }
\begin{tabular}{|c|c|c|c|c|}
\hline
Validation Accuracy \% & \multirow{2}{*}{PSPNet} & \multirow{2}{*}{DeepLabV3} & \multirow{2}{*}{Sketchgnn} & \multirow{2}{*}{\begin{tabular}[c]{@{}c@{}}EDGNet\\ (Ours)\end{tabular}} \\ \cline{1-1}
Task                   &                         &                            &                            &                                                                          \\ \hline
Text/Non-text          & 96.62                   & 96.87                      & 94.76                      & \textbf{98.48}                                                           \\ \hline
Contour/Non-contour    & 80.54                   & 82.57                      & 88.10                      & \textbf{94.57}                                                           \\ \hline
Text/Contour/Dimension & 79.54                   & 81.64                      & 84.37                      & \textbf{90.82}                                                           \\ \hline
\end{tabular}
\end{table*}

% demonstrate some sample prediction results in Fig. \ref{fig:cs_3class}. Overall our model yields much better results compared with the vision-based baseline method.

\begin{figure}[!ht]
  \centering
  \includegraphics[trim = 0in 0in 0in 0in, clip, width=\columnwidth]{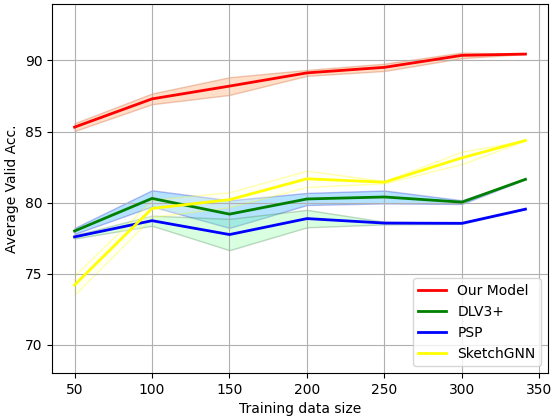}
  \caption{Validation accuracy of the models trained on datasets of various sizes.} 
  \label{fig:cs_data_size}
\end{figure}

To further compare the model stability when applied to datasets of various sizes, we conduct another set of experiments using a subset of our current train data. Each time, $k\in[50,340]$ drawings are randomly sampled from our training set for training all four models in the task of 3-class segmentation. Then we repeat the process five times with different random seeds to eliminate the influence of a biased subset taken by accident. The resulting validation accuracy curves for all four models are demonstrated in Fig. \ref{fig:cs_data_size}. It can be concluded that overall our model outperforms the other baseline models consistently by over 5\%. Another interesting finding is that graph-based methods tend to have less variance when different subsets are used compared to image-based methods, which illustrates better stability of feature extraction when encountering different data.

%% file: tex/5future.tex
With our current approach, we are able to automate the vectorization and labeling process of raster engineering drawings in terms of the separation among contour lines, dimension sets, and texts. For broader practical use in the industry, symbols such as manufacturing requirement symbols and geometric tolerance symbols are also critical in the quotation process. Considering they are usually consistent in shape and style, an immediate next step of our current work is to develop a detection algorithm to identify such symbols in the obtained vectorized results from our preprocessing step. Simple heuristic-based methods can be applied to searching for surface roughness symbols or through hole symbols since they retain highly consistent shapes (equilateral triangles) across all types of drawings. Data-driven models should be utilized to locate more complex symbols like geometric tolerance symbols because they are usually a composite of texts, symbols, and indication boxes.  

From our results on three-class segmentation, it can be concluded that contextual information about the connection between two components is needed for better accuracy. For example, edge features to indicate the angle, the shift and the curvature change between the two connected components can be added to our current graph representation. This information can help the network to understand if two components are truly connected based on semantic meaning or just independent of each other with an intersection. With this new graph representation, GraphSAGE model should also be replaced with more advanced graph networks that also take edge attributes as input for analysis, such as Graph Attention Networks \citep{velivckovic2017graph}, Graph Transformers \citep{dwivedi2020generalization}, and GINE Convolution \citep{hu2019strategies}.

The ultimate goal of our work is to aid a human operator when inspecting drawings for topology and manufacturing information necessary for the later quotation process. The effectiveness of our framework needs to be practically validated by human users. To enable easier access for the system we propose in this work, an interactive user interface should be developed to allow the users to upload their own drawings, launch the vectorization, and get the automatic component type prediction results. Then the user is only responsible for marking the critical information or correcting minor errors in the prediction. Compared to the original inspection and labeling task, the work for the human operator is more efficient and intellectual.

%% file: tex/6conclusion.tex
In this work, we present a novel framework for raster engineering drawing analysis, including a preprocessing method for drawing vectorization, a graph representation embedded with domain knowledge, and a data-driven model to learn and predict the component type of each vector in the drawing. The framework converts the problem from sparse image comprehension into semantic segmentation of the vectorized components from the original drawing, enhancing the efficiency of feature extraction. Results also show that our method yields superior performance in distinguishing the semantic meaning of contour/dimension lines compared to common CNN-based image segmentation methods. A similar framework can be established to other analyses of raster engineering drawings such as manufacturing method classification \citep{xie2022graph}, dimension estimation, and similarity search. The proposed graph representation has the potential to be used extensively in developing a digitized tool for a part quotation.